\documentclass[3p,times,procedia]{elsarticle}
\usepackage{ecrc}
\usepackage{blindtext}
\usepackage{enumitem}
\usepackage{xcolor}
\usepackage[bookmarks=false]{hyperref}
    \hypersetup{colorlinks,
      linkcolor=blue,
      citecolor=blue,
      urlcolor=blue}
\volume{00}

\firstpage{1}

\journalname{Procedia Computer Science}

\runauth{Heakl et al.}

\jid{procs}

\usepackage{amssymb}
\usepackage[figuresright]{rotating}
\usepackage{caption}
\usepackage{subcaption}
\begin{document}
\begin{frontmatter}


%% \tnotetext[label1]{Corresponding author}
\fntext[cor]{Corresponding author. Tel.: +20-1016580028; Email: ahmed.heakl@ejust.edu.eg}


\dochead{6th International Conference on AI in Computational Linguistics}%

\title{ResuméAtlas: Revisiting Resume Classification with Large-Scale Datasets and Large Language Models}

\author[a,*]{Ahmed Heakl\fnref{cor}} 
\author[a,*]{Youssef Mohamed}
\author[a,*]{Noran Mohamed}
\author[a]{Aly Elsharkawy}
\author[a]{Ahmed Zaky}

\address[a]{Egypt-Japan University of Science and Technology, New Borg El-Arab City, 21934, Alexandria, Egypt}
\address[*]{Equal Contribution}

\begin{abstract}
The increasing reliance on online recruitment platforms coupled with the adoption of AI technologies has highlighted the critical need for efficient resume classification methods. However, challenges such as small datasets, lack of standardized resume templates, and privacy concerns hinder the accuracy and effectiveness of existing classification models. In this work, we address these challenges by presenting a comprehensive approach to resume classification. We curated a large-scale dataset of 13,389 resumes from diverse sources and employed Large Language Models (LLMs) such as BERT and Gemma1.1 2B for classification. Our results demonstrate significant improvements over traditional machine learning approaches, with our best model achieving a top-1 accuracy of 92\% and a top-5 accuracy of 97.5\%. These findings underscore the importance of dataset quality and advanced model architectures in enhancing the accuracy and robustness of resume classification systems, thus advancing the field of online recruitment practices. Our models, code, and dataset are available as open-source resources. \footnote{Code: \url{https://github.com/noran-mohamed/Resume-Classification-Dataset}} \footnote{Models: \url{https://huggingface.co/collections/ahmedheakl/resumeatlas-668047e86bc332049afd0b39}} \footnote{Dataset: \url{https://huggingface.co/datasets/ahmedheakl/resume-atlas}}
\end{abstract}

\begin{keyword}
Resume Classification; Online Recruitment; Dataset Collection; Large Language Models (LLMs); Transformers




\end{keyword}

% \cortext[cor1]{Corresponding author. Tel.: +20-1016580028; Email: ahmed.heakl@ejust.edu.eg}
\end{frontmatter}
% \email{ahmed.heakl@ejust.edu.eg}

% \linenumbers


%\enlargethispage{-7mm}
\section{Introduction}
In today's fast-paced and competitive job market, online recruitment has become integral to employers and job seekers. With the global online recruitment market projected to reach USD 39.76 billion in 2022 and expected to grow at a Compound Annual Growth Rate (CAGR) of 7.2\% between 2021-2026, understanding the intricacies of this evolving landscape is paramount \cite{rec_stat}. One crucial aspect of online recruitment that demands attention is resume classification \cite{Bil}. As nearly 70\% of companies utilize online recruiting platforms and 94\% of employers plan to adopt or continue using AI for talent acquisition and HR, the volume of job applications is skyrocketing \cite{dg}. Consequently, efficiently categorizing and analyzing resumes has emerged as a critical challenge for recruiters. 

Despite the growing reliance on online recruitment platforms and the adoption of AI technologies, several challenges persist in resume classification. One significant hurdle is the availability and quality of datasets for training AI algorithms. The collection of datasets from recruiting platforms poses challenges due to the sensitive nature of the data, which often contains personal information about job seekers \cite{carrot}. Ensuring data privacy and compliance with regulations such as GDPR (General Data Protection Regulation) adds complexity to the dataset collection process. Additionally, the lack of standardized templates for resumes further complicates resume classification efforts. With no universal guidelines for what constitutes a high-quality resume, recruiters often encounter a plethora of resumes varying in format, structure, and content quality \cite{pr}. The variability in resume quality poses challenges for AI algorithms, potentially leading to biases and inaccuracies in candidate selection. \cite{var}. Addressing these challenges is crucial for advancing resume classification and improving online recruitment efficiency. 

A prevalent problem in the current state-of-the-art of resume classification is the reliance on small datasets with limited samples and labels. Many researchers and practitioners in the field often work with datasets containing only a few thousand samples and a small number of labels, typically ranging from 5 to 25 categories. This scarcity of data presents significant challenges for training robust and accurate classification models. Moreover, the use of classic machine learning algorithms such as Naïve Bayes \cite{nv}, Support Vector Machine (SVM) \cite{svm}, Random Forest \cite{randomforest}, K-Nearest Neighbor \cite{KNN}, and Logistic Regression \cite{lr}, alongside TF-IDF vectorization \cite{tf} and XGB algorithms \cite{xgb}, exacerbates the issue. While these algorithms have proven effective in various contexts, their performance may be suboptimal when confronted with large datasets and complex classification tasks \cite{small}, leading to challenges in achieving high accuracy and generalization. As such, overcoming the limitations imposed by small datasets and exploring innovative approaches to improve model performance remains a critical area of research in the field of resume classification.

The contributions of this work represent a significant advancement in the field of resume classification, addressing key challenges and pushing the boundaries of existing methodologies. Our contributions are as follows:

\begin{itemize}
    \item \textbf{Large-Scale Dataset Collection and Preprocessing} The curated dataset, comprising 13,389 records from diverse sources across 43 distinct resume categories, represents the largest collection for resume classification to the best of our knowledge, underpinned by approximately 400 hours of meticulous data preprocessing efforts aimed at ensuring high-quality samples and minimizing noise and inconsistencies, thereby enhancing the reliability and robustness of the model.
    \item \textbf{Outperforming State-of-the-Art Models} We employed cutting-edge LLMs, including Gemini and BERT, to achieve remarkable accuracy, with 91\% top-1 accuracy and 97\% top-5 accuracy, surpassing existing state-of-the-art models in resume classification tasks. 
    \item \textbf{Providing High-Quality Codebase} We provide high-quality codes for scraping, preprocessing, and training, facilitating reproducibility and enabling researchers to build upon our work. 
\end{itemize}

To overcome limitations in resume classification, we took a comprehensive approach to dataset collection and model development. We gathered a high-quality dataset from diverse sources, including Google, Bing, and leading resume websites. This endeavor involved meticulous data collection and filtering processes, totaling approximately 400 hours, to ensure the inclusion of diverse samples representing a wide range of resume categories. As a result, we assembled the largest dataset in the context of resume classification, comprising 13,389 records across 43 classes. Leveraging state-of-the-art transformer models and Large Language Models (LLMs) such as Gemma \cite{Gemma} and BERT \cite{BERT}, we tackled the complexities of resume classification with a focus on achieving robustness and high accuracy. By employing advanced techniques and leveraging the richness of the collected dataset, our approach transcended the limitations of traditional machine learning models, resulting in the development of a highly effective and reliable model that outperforms the shortcomings of existing state-of-the-art models. 

\par
The rest of this paper is presented as follows: Section \ref{sec:related-works} reviews related works in the field of resume classification. Section \ref{sec:dataset} describes our dataset collection process, challenges, and preprocessing steps. Section \ref{sec:methods} outlines the methodologies employed in our study, including the use of LLMs and traditional machine learning approaches. Section \ref{sec:results} presents and discusses our experimental results. Finally, Section \ref{sec:conclusion-future-work} concludes the paper and suggests directions for future research.

\section{Related Works}\label{sec:related-works}
Pal et al. (2022), \cite{rc-various-ml} addressed the imperative for automating the hiring process amid the transition to remote work precipitated by the epidemic. It confronts the central issue of inefficiency and manual effort inherent in resume categorization and verification, proposing automation as a substantial alleviating measure. Leveraging machine-learning algorithms, including Naïve Bayes, SVMs, and random forest, the investigation endeavors to extract skills and categorize resumes into pertinent job profile classes. Data acquisition spans various sources such as Kaggle, Glassdoor, and Indeed, yielding unstructured datasets subjected to cleaning, classification, and storage processes. A training dataset encompassing 70\% of the total dataset is utilized. Findings reveal the superiority of the random forest model over Naïve Bayes and SVM, yielding an accuracy rate of 70\% and exhibiting enhanced predictive capabilities. 

Skondras et al. (2023) \cite{rc-dnn} explored using synthetic resumes to augment training data and improve the effectiveness of resume classification algorithms, particularly in categories with sparse samples. The authors employed the OpenAI API to generate structured and unstructured resumes and resumes from the Indeed website with total records of 4791 in the ChatGPT Dataset and 1533 in the Indeed Dataset with 15 categories. These synthetic resumes were used to train two models: a transformer model (BERT) and a feedforward neural network (FFNN) incorporating Universal Sentence Encoder 4 (USE4) embeddings. The experiments showcased that the BERT mode, coupled with augmented datasets, demonstrated superior performance compared to the FFNN model. A 92\% accuracy was achieved in the sixth experiment, where the BERT model was combined with the Indeed augmented dataset. 

Additionally, Ali et al. (2022) \cite{rc-nlp} presented an NLP-based approach to classify resumes into job categories. The study employs a dataset of 962 labeled resumes across 25 job categories and evaluates nine ML classification models, including SVM, Naïve Bayes, and Logistic Regression. The results show that SVM classifiers, particularly the Linear Support Vector Classifier, achieve an accuracy of over 96\%. 

Jalili et al. (2024) \cite{Bil} presents a novel method for resume classification using a Bidirectional LSTM (BiLSTM) architecture to enhance the accuracy and efficiency of candidate evaluation in the recruitment process. Their method includes text preprocessing steps, followed by utilizing BiLSTM to capture both past and future contexts of resume content. Word embedding is used to enrich the textual representations. The dataset is obtained from LiveCareer and comprises a collection of over 2400 resumes, categorized into 21 distinct job categories. The BiLSTM is good at capturing sequential dependencies achieving 72.4\% in classification accuracy.

Nasser et al. (2018) \cite{rc-cnn} concentrated on the document classification domain, specifically focusing on resume categorization into distinct classes. The proposed methodology utilizes Convolutional Neural Networks (CNN) with Glove-Word Embedding for resume classification. Resumes undergo hierarchical segmentation, and a CNN model with word embedding is employed at each level for classification. The outputs from individual classifiers are amalgamated to define the overarching resume category hierarchy. 
The study utilizes datasets from Calpine Lab's resume collection and sample job descriptions covering both technical and non-technical domains. Eight classifiers are used for the classification task, covering binary and multi-class classifiers. The classification hierarchy spans five levels, with each level tailored to specific tasks and employing dedicated CNN models. The dataset includes multiple labels across 20 categories distributed across hierarchy levels. The CNN architecture comprises embedding, convolutional, max-pooling, dropout, and dense layers, and the system's efficacy is assessed using precision, recall, and f-score metrics. Results demonstrate promising performance, with high accuracy levels observed across different hierarchical levels. For instance, at Level 1, the training accuracy attained 99\%, while the test accuracy achieved 94\%. Similarly, at Level 5, the training accuracy recorded 98.7\%, and the test accuracy reached 92.9\%. These findings underscore the effectiveness of the CNN-based approach in accurately classifying resumes into specific categories, thereby facilitating the recruitment and selection processes of candidates.

Within the online job recruitment domain, the precise categorization of job postings and resumes holds significant significance for both job seekers and recruiters alike. Ramraj et al. (2020) \cite{rc-linkedin} introduced an automated text classification system tailored to classify textual data, specifically resumes, into diverse categories utilizing advanced methodologies such as Term Frequency-Inverse Document Frequency (TF-IDF) with CNNs. The dataset comprises over 1000 resumes obtained from LinkedIn through web scraping techniques and API tools. Each resume includes attributes like job title, description, skills, location, and past experiences. Resumes are segmented into multiple labels based on job titles and descriptions, resulting in a dataset with 27 categories. Domain adaptation techniques are used due to the sensitive nature of resume data. Preprocessing ensures data quality and consistency, with resumes classified into multiple labels based on job titles and descriptions. The CNN algorithm, typically used for image classification, is adapted to extract character-level features from LinkedIn profiles. Various algorithms, including SVM, Naive Bayes (NB), and TF-IDF, are applied individually and collectively to the dataset. Outcomes reveal the CNN algorithm's superiority, manifesting in an accuracy rate of 68\% and the highest F1-score of 0.65 among all evaluated models. Additionally, TF-IDF coupled with NB, SVM, and XGB algorithms also demonstrates competitive performance, yielding F1-scores ranging from 0.57 to 0.61. This study contributes valuable insights into the effectiveness of different algorithms for resume classification, offering implications for improving online job recruitment processes. 

\begin{figure}[t]
    \centering
    \begin{subfigure}[b]{0.3\textwidth}
      \includegraphics[width=\linewidth]{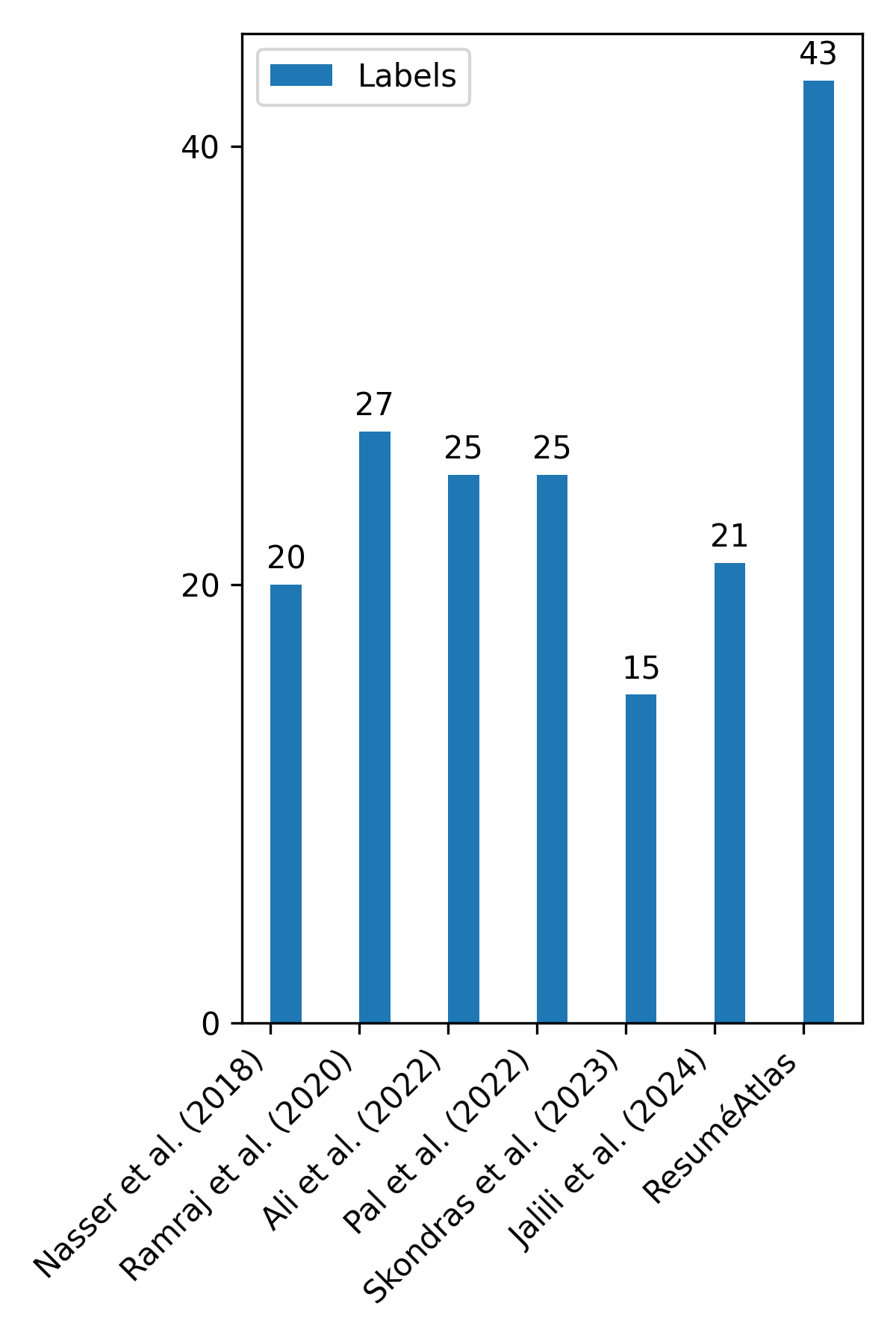}\par
      \caption{Number of the labels in each dataset.}
      \label{fig:num-labels} 
    \end{subfigure}
    % ------------------------------- %
    \begin{subfigure}[b]{0.3\textwidth}
      \includegraphics[width=\linewidth]{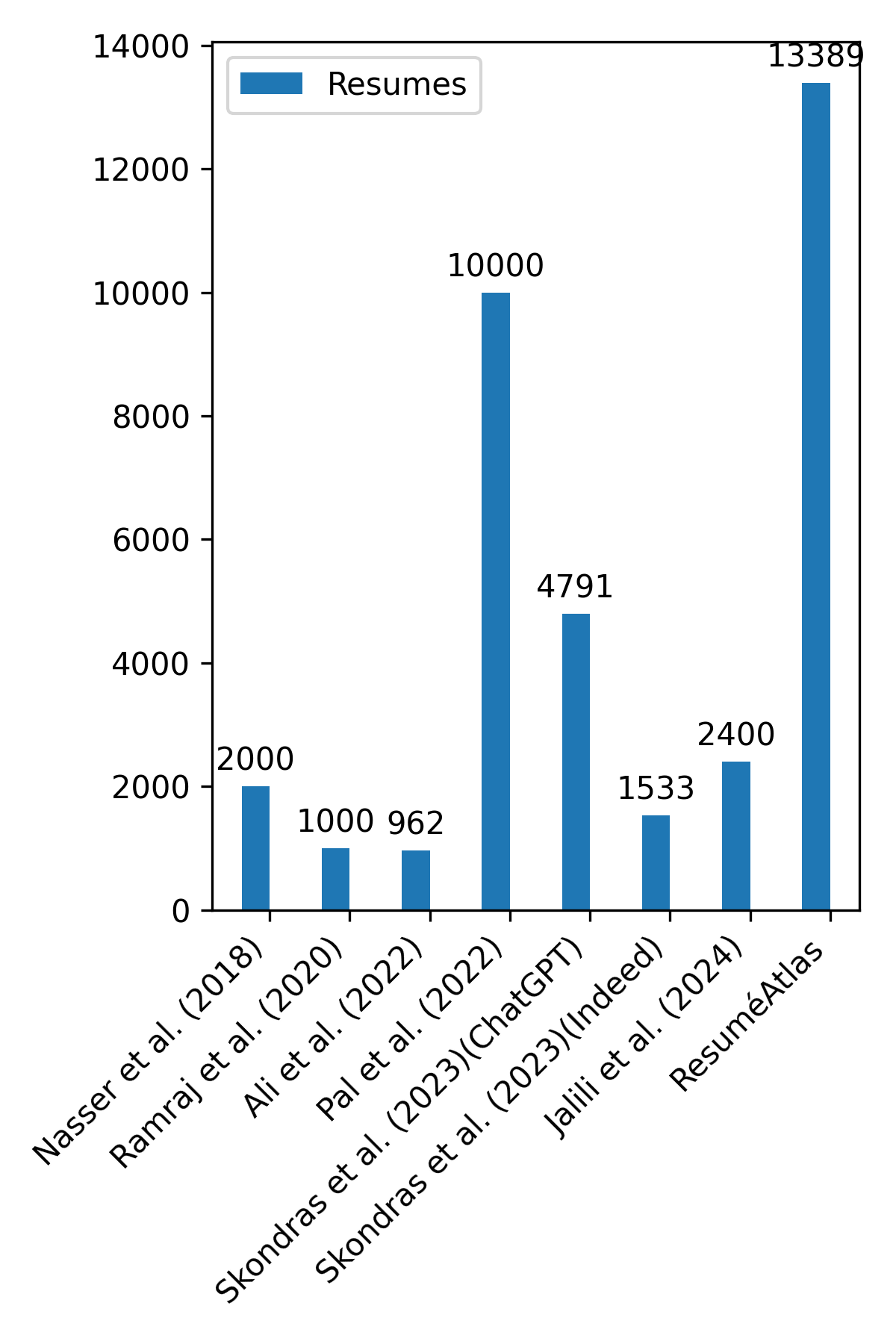}\par
      \caption{Number of the collected resumes.}
      \label{fig:num-resumes} 
    \end{subfigure}
    \caption{Collected dataset samples and labels comparison with other datasets.}
\end{figure}

\section{Dataset}\label{sec:dataset}
We collected "ResuméAtlas", a dataset comprised of resumes sourced from Google Images, Bing Images, and LiveCareer, with textual content extracted using optical character recognition (OCR). A total of 13,389 records were obtained through automated scraping techniques, including 3,015 from Google Images, 2,722 from Bing Images, and 7,652 from LiveCareer. A comparison of the number of collected resumes and labels in this dataset to others is presented in figures \ref{fig:num-labels} and \ref{fig:num-resumes}.

\subsection{Data Collection Process}
The collection process involved several stages, starting with the scraping of resume images from each source. This process was executed using separate scripts tailored for Google Images, Bing Images, and LiveCareer. The scraping phase required approximately 5 hours for Bing, about 25 hours for Google, and nearly 40 hours for LiveCareer due to the websites' different complexities and the data volume. Following the scraping phase, the downloaded images underwent extensive filtering procedures over approximately 80 hours to ensure the quality and relevance of the data by iterating over the downloaded resumes one by one to delete the irrelevant ones. In the final stage of data preprocessing, Optical Character Recognition (OCR) algorithms were employed to extract textual content from the resume images. This process was performed separately for the Google/Bing and LiveCareer datasets, requiring approximately 95 hours and 145 hours, respectively, to process the entire dataset. Specifically, Google's Cloud Vision service was utilized to facilitate the OCR process.

\subsection{Challenges in the Dataset}
The dataset presents several challenges that must be addressed during analysis. These challenges include the lack of a structured format, making it difficult to extract specific information such as name, education, and experience. Additionally, resumes may contain personal information such as names, phone numbers, and email addresses scattered throughout the text, posing privacy concerns and necessitating the implementation of data anonymization techniques. Skills or experiences may be mentioned multiple times in different sections of the resume, leading to redundancy and requiring techniques to identify and remove duplicates. Misspellings, such as "experiance" instead of "experience" or "technicl" instead of "technical", may be present in the text, affecting the quality of the data and necessitating the implementation of spell-checking algorithms. Headers, footers, and contact information at the top or bottom of the resume may not be relevant to the content analysis. They should be filtered out to prevent noise and improve data quality. Some resumes may contain watermarks or highlighted sections that need to be removed or accounted for during analysis to prevent bias and ensure accurate results. Special characters such as '*', '/', '\&', '\$', '\%', '$\wedge$', '$\sim$' may be present in the resume text and need to be handled appropriately during preprocessing. Resumes may also contain URLs or links to personal websites or online profiles. Finally, resumes may contain experience irrelevant to the job being applied for, which might need to be identified and filtered out during analysis to improve the accuracy of the results. These challenges highlight the importance of developing robust preprocessing and analysis techniques to ensure the quality and accuracy of the results.

\subsection{Data Preprocessing}
In the data preprocessing phase, we employed a series of steps to ensure the cleanliness and uniformity of the text data. We began by converting all text to lowercase to promote consistency. Next, we utilized regular expressions to remove punctuation marks and non-alphanumeric characters systematically. We also removed URLs, Twitter handles, hashtags, and special characters, and expanded contractions. Finally, we removed common stop words, such as 'and', 'the', and 'is', using the NLTK \cite{nltk} stop words corpus, thereby enhancing the quality of the tokenized text.

\section{Methods}\label{sec:methods}
Resume classification is a fundamental task in natural language processing, where the goal is to assign a relevant class label to a given text input $c_{pred} = \arg\max_{c_{i} \in C} P(c_{i} | w_0, w_1, \ldots, w_n, \theta)$ where $c_{pred}$ represents the predicted class label, $C$ denotes the set of possible class labels, and $P(c_{i} | w_0, w_1, \ldots, w_n)$ represents the probability of the $i^{th}$ class label given the input text, which is composed of words $w_0, w_1, \ldots, w_n$. Here $\theta$ represents the model weights which is optimized through training.

In this study, we leveraged the sequential nature of the resume classification problem to employ large language models (LLMs) as our classification approach. Specifically, we explored two prominent variants of LLMs: Bidirectional Encoder Representations from Transformers (BERT) and Gemma1.1 2B \cite{gemma}, \cite{bert}.

For the BERT-based approach, we utilized an encoder-based architecture that receives a sequence of words as input, which are then converted into tokens and embedded. Positional embeddings were added to the textual embeddings to preserve sequential information. The model employs multi-head self-attention blocks to extract meaningful representations from the input sentences. The output of these stacked attention blocks is a matrix of shape $(b, t, d)$, where $b$ is the batch size, $t$ is the sequence length, and $d$ is the embedding dimension. These extracted features were then fed into a feedforward neural network, which outputs a vector out of dimension $(b, c)$, where $c$ is the number of classes. The predicted class label was determined by taking the $argmax$ of the output vector.

In addition to BERT, we also explored the use of Gemma1.1 2B, a decoder-based architecture. This model was trained using an auto-aggressive approach with an attention mask on the output classification.

To compare with \cite{rc-various-ml, rc-dnn, rc-nlp, rc-jrc, rc-cnn, rc-linkedin}, to utilize machine learning approaches, we used Term Frequency - Inverse Document Frequency (TF-IDF) as a feature extractor. TF-IDF is a statistical weighting technique that evaluates the importance of a term within a document by considering its frequency of occurrence, while also taking into account its rarity across the entire corpus. By balancing term frequency and inverse document frequency, TF-IDF provides a robust method for feature extraction and dimensionality reduction in text analysis. The extracted features from TF-IDF were inputted into Support Vector Machines (SVM) \cite{svm}, logisitc regression \cite{lr}, XGBoost \cite{xgb}, multi-layer perceptron (MLP), naive bayes multinomial \cite{naive-bayes}, and random forest. 

The dataset, comprising 13,389 samples, was divided into three subsets: 2,677 samples (20\%) for testing, 1,071 samples (10\%) for validation, and 9,640 samples (70\%) for training. The training process for BERT and Gemma models was conducted on 2xT4 16GB GPUs. Specifically, BERT was trained for 7 epochs, while Gemma was trained for 1 epoch. In contrast, the classical methods (TF-IDF + $<$method$>$) were trained using the sci-kit learn library \cite{sci-kit} on a Ryzen 7 12-core CPU. Notably, the training process for Gemma involved quantization-aware training with fp16 precision. Additionally, we employed LORA \cite{lora} to fit model weights on the GPU. The optimization process for both BERT and Gemma utilized AdamW \cite{adamw} with a cosine learning rate scheduler. Furthermore, we implemented gradient check pointing \cite{gradient-checkpoint} and gradient accumulation \cite{gradient-accumulation} to reduce GPU requirements, as the Adam optimizer maintains multiple copies of the weights.

\section{Results \& Discussion}\label{sec:results}
\begin{table}[t] 
    \centering 
    \begin{tabular}{|l|c|c|c|c|} 
        \hline \textbf{Method Name} & \textbf{Accuracy (\%)} & \textbf{Precision (\%)} & \textbf{Recall (\%)} & \textbf{F1 Score (\%)} \\ \hline 
        TF-IDF + Random Forest \cite{rc-jrc} & 78.5 & 75.2 & 82.1 & 78.6 \\  
        TF-IDF + SVM \cite{rc-various-ml, rc-nlp, rc-linkedin} & 79.2 & 76.5 & 83.5 & 80.0 \\  
        TF-IDF + Logistic Regression \cite{rc-nlp} & 79.8 & 77.1 & 84.2 & 80.6 \\ 
        TF-IDF + Naive Bayes Multinomial \cite{rc-various-ml, rc-nlp, rc-linkedin} & 81.3 & 79.5 & 86.2 & 82.8 \\  
        TF-IDF + MLP \cite{rc-dnn} & 81.6 & 80.2 & 87.1 & 83.6 \\  
        BiLSTM \cite{Bil} & 81.8 & 78.3 & 83.5 & 79.8 \\
        TF-IDF + XGB \cite{rc-linkedin} & 83.5 & 82.1 & 89.5 & 85.8 \\ \hline
        BERT & 91.2 & 90.5 & 92.8 & 91.6 \\ 
        BERT - Top 3 & 96.1 & 95.8 & 97.3 & 96.5 \\ 
        BERT - Top 5 & 97.5 & 97.2 & 98.1 & 97.6 \\  
        BERT - Top 10 & 98.5 & 98.3 & 99.1 & 98.7 \\  
        Gemma1.1 2B & \textbf{92.0} & \textbf{91.5} & \textbf{93.2} & \textbf{92.3} \\ \hline 
    \end{tabular} 
    \caption{Comparison of classification accuracy, precision, recall and f1 scores for different methods. The lower section presents our work.} 
    \label{tab:accuracy} 
\end{table}
We present the accuracy, precision, recall, and f1-score of each model on the test dataset, comprising 2,678 samples. For BERT, we report top-1, top-3, top-5, and top-10 accuracy, as some resumes can have multiple valid job titles. This is reflected in the top-x suggested titles according to a pre-chosen top-p probability.

As shown in Table \ref{tab:accuracy}, attention-based models (Gemma, BERT) outperform their classical machine learning counterparts. The top-performing transformer-based model, Gemma1.1 2B, surpasses the top-performing classical model, TF-IDF + XGB, by 8.5\%. This can be attributed to the ability of attention-based architectures to leverage the auto-aggressive nature of text, unlike TF-IDF, which relies solely on word frequency. Additionally, XGBoost performs well among classical machine learning methods, owing to its ensemble learning and regularization capabilities. Moreover, TF-IDF produces complex features that require a large space to represent, making MLP the best performer among classical methods, except for XGBoost.

\begin{figure}[t]
    \centering
    \begin{subfigure}[b]{0.60\textwidth}
      \includegraphics[width=\linewidth]{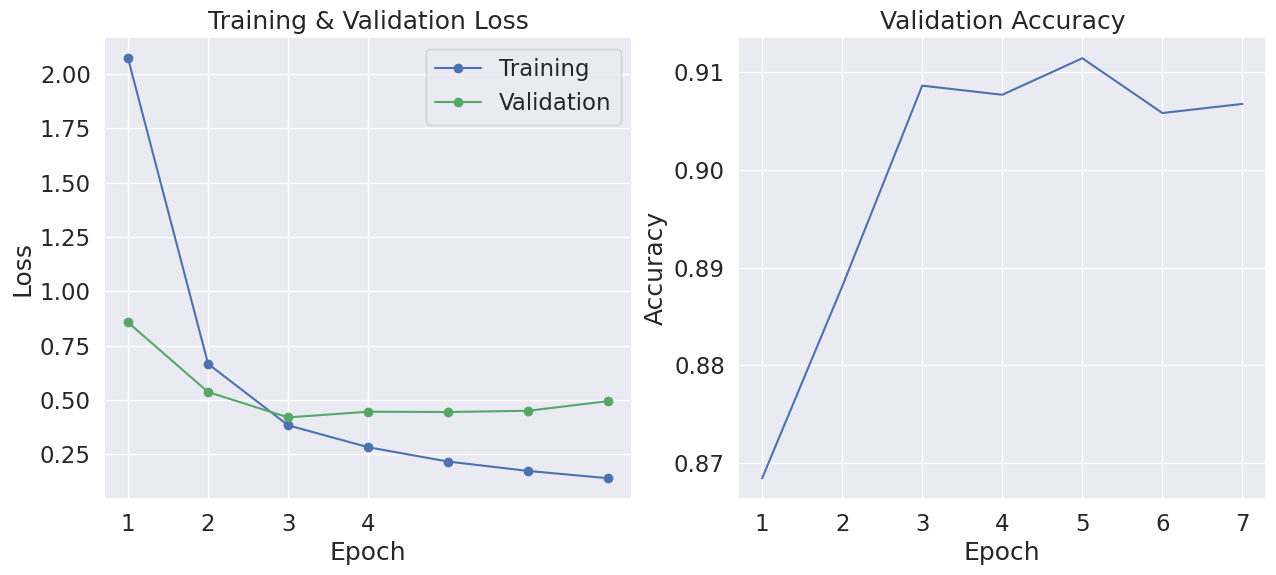}\par
      \caption{BERT training loss and accuracy.}
      \label{fig:bert-loss} 
    \end{subfigure}
    \hfill
    % ------------------------------- %
    \begin{subfigure}[b]{0.39\textwidth}
      \includegraphics[width=\linewidth]{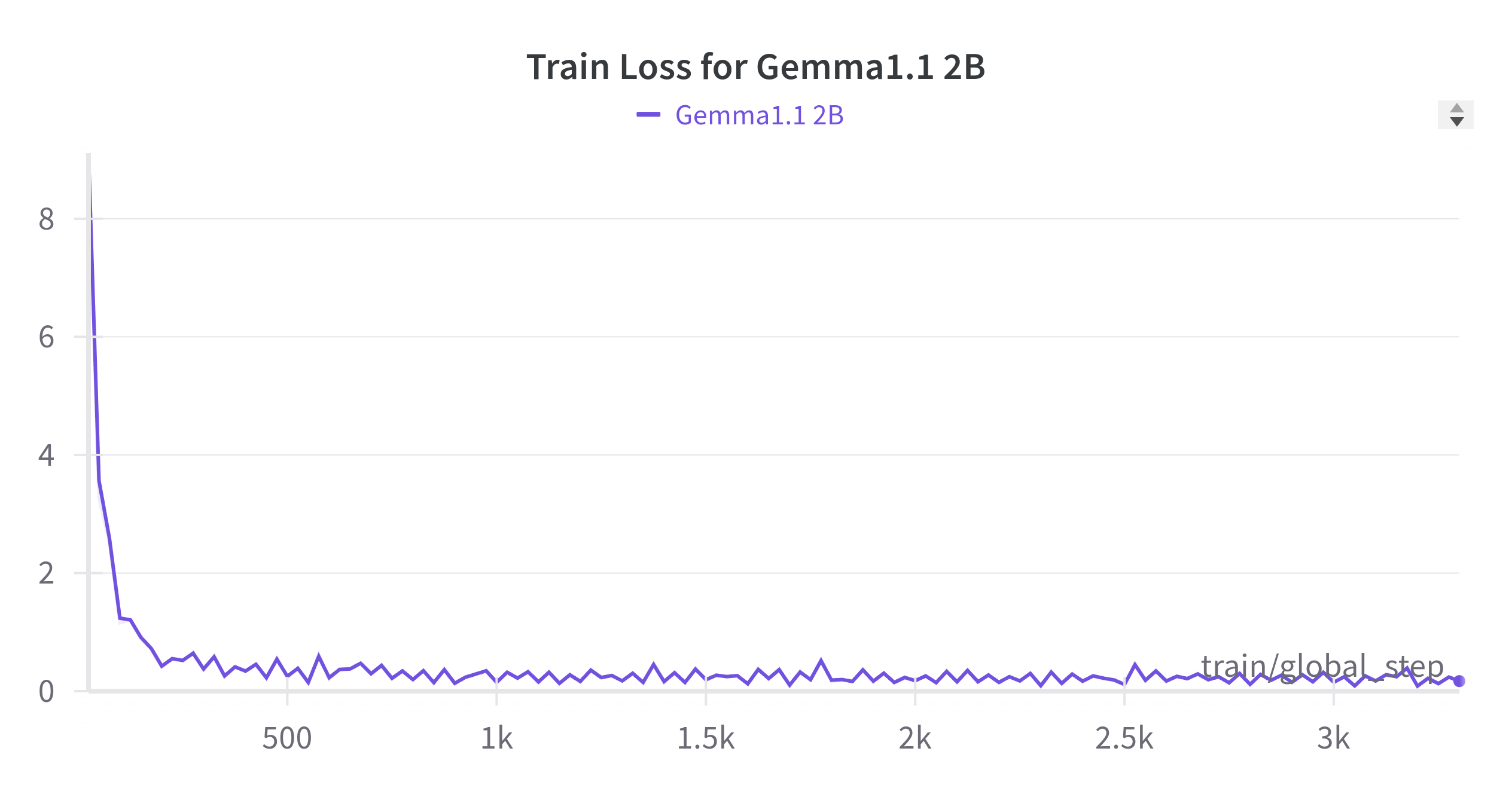}\par
      \caption{Gemma training loss.}
      \label{fig:gemma-loss} 
    \end{subfigure}
    \hfill
    % ------------------------------- %
    \begin{subfigure}[b]{0.4\textwidth}
      \includegraphics[width=\linewidth]{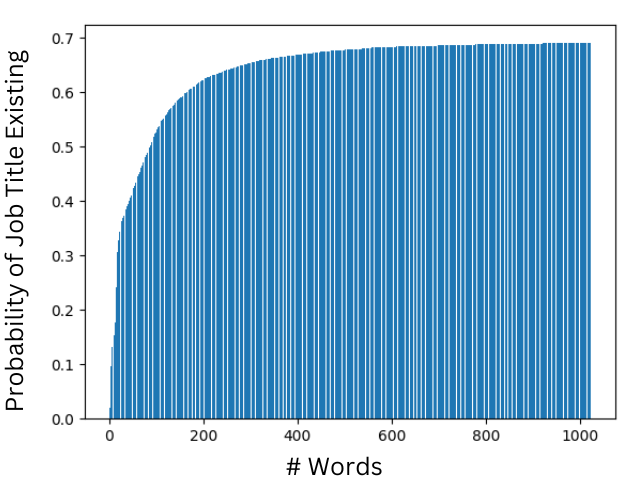}\par
      \caption{Probability of title in text vs. the number of words in text.}
      \label{fig:title-per-words} 
    \end{subfigure}
    % ------------------------------- %
    \begin{subfigure}[b]{0.4\textwidth}
      \includegraphics[width=\linewidth]{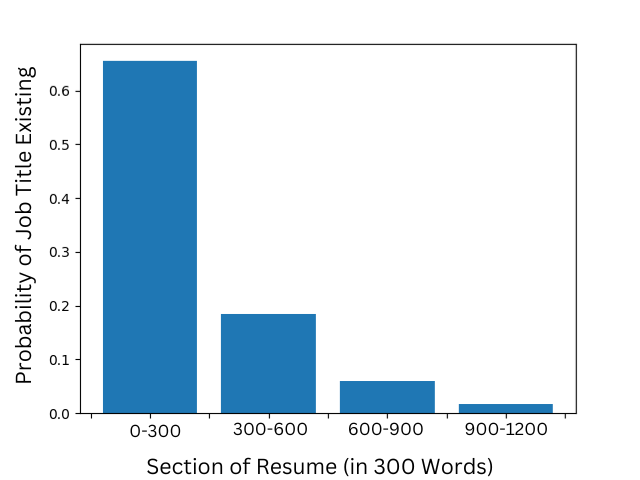}\par
      \caption{Probability of title in text vs. 300-word sections.}
      \label{fig:prob-per-section} 
    \end{subfigure}
    % \caption{Training curves and dataset analysis.}
\end{figure}

Notably, all models were trained on only the first 300 words of the text. As illustrated in Figures \ref{fig:title-per-words}, \ref{fig:prob-per-section}, nearly 70\% of job titles are present in the first 300-500 words, and the probability of finding the job title in the text does not increase significantly beyond 500 words. This suggests that important information is typically found at the head of the resume.

Figure \ref{fig:gemma-loss} shows that Gemma training saturates quickly, around 250 steps out of 3900 steps (only 6\%). This highlights the power of large language models in solving text classification tasks, particularly decoder-based architectures that exploit the auto-aggressive nature of input text. In contrast, Figure \ref{fig:bert-loss} reveals that, after the third epoch, the training loss decreases while the validation loss increases, indicating overfitting behavior in multi-epoch training for encoder-based architectures.

As evident from Table \ref{tab:accuracy}, increasing the number of output in BERT from top-1 to top-3 yields a 5.4\% accuracy improvement, supporting our claim that some resumes can be appropriately labeled with multiple titles. This may be due to individuals holding multiple jobs or a single job being attributed to multiple titles. For example, the occupation of software engineer can be interchangeably referred to as frontend engineer, backend engineer, Python developer, full-stack engineer, and other related titles. Hence, considering the top-x samples would make sense.

Finally, the high accuracy of our models can be attributed to the large size of our dataset, comprising 13,000 examples, and the diversity of job titles, with nearly twice as many titles as the largest existing dataset (Figure \ref{fig:title-per-words}). This ensures our models generalize well to unseen labels.
\section{Conclusion and Future Work}\label{sec:conclusion-future-work}

In this study, we addressed significant challenges in resume classification within the online recruitment domain by leveraging a large-scale dataset and advanced Language Model Models (LLMs). Through meticulous dataset collection efforts, we assembled a comprehensive dataset of 13,389 resumes across 43 distinct categories, overcoming limitations associated with small datasets and resume format variability. By employing state-of-the-art transformer models such as BERT and Gemma1.1 2B, we achieved remarkable improvements in classification accuracy, with top-performing models reaching \textbf{92\%} top-1 accuracy and \textbf{97.5\%} top-5 accuracy, outperforming traditional machine learning approaches.

Future research directions include expanding the dataset with diverse sources, broadening the scope of job titles, and exploring techniques for handling resume variability. Investigating domain-specific pre-training for LLMs and developing methods to interpret model decisions could further enhance the applicability and transparency of resume classification in real-world recruitment scenarios.

\end{document}